# Implementation of Neural Network and feature extraction to classify ECG signals

R karthik, Dhruv Tyagi, Amogh Raut, Soumya Saxena, Rajesh Kumar M, Senior IEEE Member

Department of Electronics and Communication Engineering VIT University Vellore, India 632014
tkgravikarthik@gmail.com, mrajeshkumar@vit.ac.in

Abstract: This paper presents a suitable and efficient implementation of a feature extraction algorithm (Pan Tompkins algorithm) on electrocardiography (ECG) signals, for detection and classification of four cardiac diseases: Sleep Apnea, Arrhythmia, Supraventricular Arrhythmia and Long Term Atrial Fibrillation (AF) and differentiating them from the normal heart beat by using pan Tompkins RR detection followed by feature extraction for classification purpose .The paper also presents a new approach towards signal classification using the existing neural networks classifiers.

# 1 Introduction

Electrocardiography (ECG) is a technique used to record electrical activity of the heart and observe the heart variation and abnormalities over a period of time using electrodes placed on the skin. ECG signal can be divided into phases of depolarization and repolarization of the muscle fibers which make up the heart [9].These phases consist of P-waves, QRS-complexes and T-waves and provide fundamental information about the electrical activities of the heart [2]. ECG signal processing can be used to detect diseases including Arrhythmia , Supraventricular Arrhythmia  Sleep Apnea, Normal Sinus Rhythm  and Long Term Atrial fibrillation[1]. Sleep apnea is a sleep disorder characterized by cessations of breathing during sleep. There are three types of sleep apnea: Central Sleep Apnea (CSA), Obstructive Sleep Apnea (OSA) and mixed . Arrhythmia occurs due to factors like Coronary artery disease, Electrolyte imbalances in blood, Changes in heart muscle etc**.** Supraventricular Arrhythmia is a type of arrhythmia causing an abnormally fast heart rhythm due to unsuitable electrical activity in the heart. It begins in the areas above the heart's lower chambers, such as the upper chambers (the atria) or the atrial conduction pathways. This disorder can result from rheumatic heart disease or an overactive thyroid gland. Long Term Atrial fibrillation (AF) involves occurrence of an irregular heartbeat where the atria fails to contract in a strong manner. The clinical risk factors for AF include advancing age, diabetes, hypertension, congestive heart failure, rheumatic and non-rheumatic valve disease, and myocardial infarction. The echocardiographic risk factors for non-rheumatic AF includes left atrial enlargement, increased left ventricular wall thickness, and reduced left ventricular fractional shortening. ECG signals available from Physionet library provide a standard dataset for performing all tests. ECG signal processing is used in order to convert the raw data into a form which can be used for feature extraction.

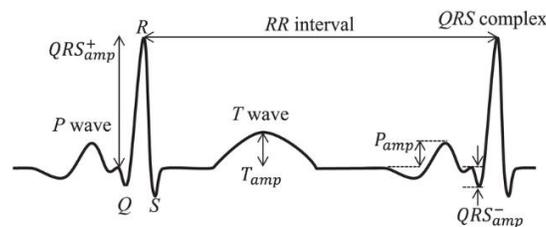

Fig 1. Normal ECG Heart Beat [14]

Discrete Wavelet Transform provides a method for feature extraction where the choice of the wavelet selected lies upon the application and the user. Wavelet families include Biorthogonal, Coiflet, Harr, Symmlet, Daubechis [15] wavelets [11] [5]. Wavelet techniques are the most commonly used but is complex and time consuming. Hence, other techniques like Pan Tompkins algorithm should be used for preprocessing and feature extraction as it provides a higher level of decomposition and comparatively less time consuming [12][13]. Fast Fourier transform and other techniques are used for pre-processing of the signal in order to remove noise and baseline wandering [10].Several classification techniques can be used for ECG classification including Support Vector Machines (SVM), decision tree, neural network, nearest neighbors, etc [6]. Linear discriminant analysis is a linear classifier that minimizes the interclass variance and maximizes the mean values of the two classes to find a line in lower dimension of feature space [3]. They do not take into account the difference between adjacent sample points. Support Vector Machines (SVM) on the other hand use the adjacent sample points to draw a discriminatory line used for classification [3]. SVM is considered to give higher accuracies and hence is preferable. Artificial Neural Networks. ANN classifiers can be fed by various parameters three of which can be spectral entropy, Poincare plot geometry, and largest Lyapunov exponent (LPE) [9].

**Feature extraction**

The raw ECG signal is processed to filter out noise and extract the RR interval using Pan Tompkins algorithm [13] which is further used to extract 15 features out of each signal. The extracted features are fed into 4 different neural networks for training and are then validated using various test files. The accuracy is calculated for each neural network and each disease. The same process is performed for the proposed classification technique as well and the results are compared. ECG signal includes noise as a part of the signal which needs to be removed before processing is done on it for feature extraction. Pan-Tompkins algorithm is a real-time algorithm for detection of the QRS complexes of ECG signals developed by Jiapu Pan and Willis J. Tompkins [13]. It reliably recognizes QRS complexes on the basis of digital analysis of slope, amplitude, and width. In this algorithm, a special digital band pass filter reduces false detections which are caused by the various types of interferences present in ECG signals. This filtering allows the use of low thresholds, and hence helps in increasing the detection sensitivity. Stepwise signal processing of the raw signal of each disease is graphically depicted in Fig. 9-13.

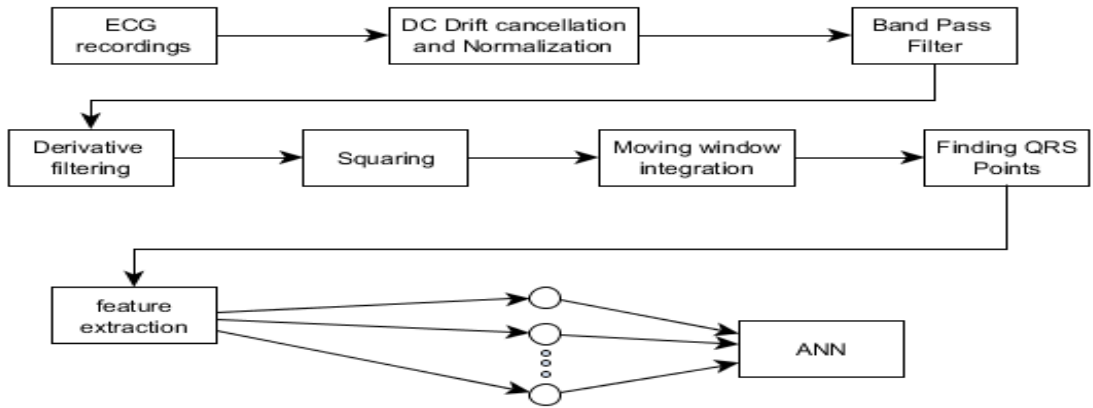

Fig. 8. Block Diagram for Pan Tompkins algorithm (gray dotted box) to derive features used as ANN inputs

Feature extraction was done using Heart Rate Variability (HRV) signals. HRV can be defined as the interval between successive R peaks.
$$rr(i) = r(i+1) - r(i); 1,2,\ldots,m-1 \tag{1}$$
where r(i) is the R peak time for i[th] wave [1].
The extracted RR interval from each data segment was used to extract a total of 15 features [1]

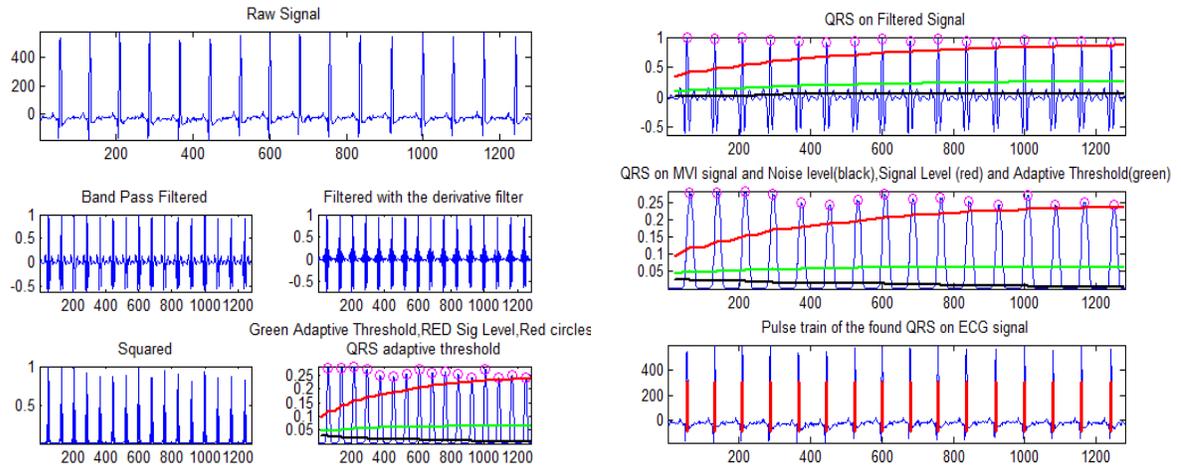

Fig.9. Signal Processing of Normal Raw ECG signal

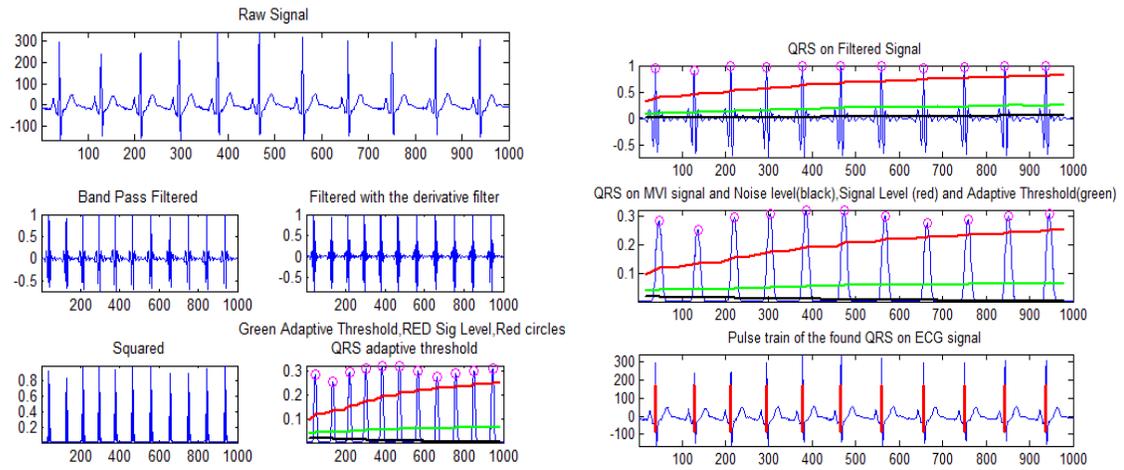

Fig.10. Signal Processing of Sleep Apnea Raw ECG Signal

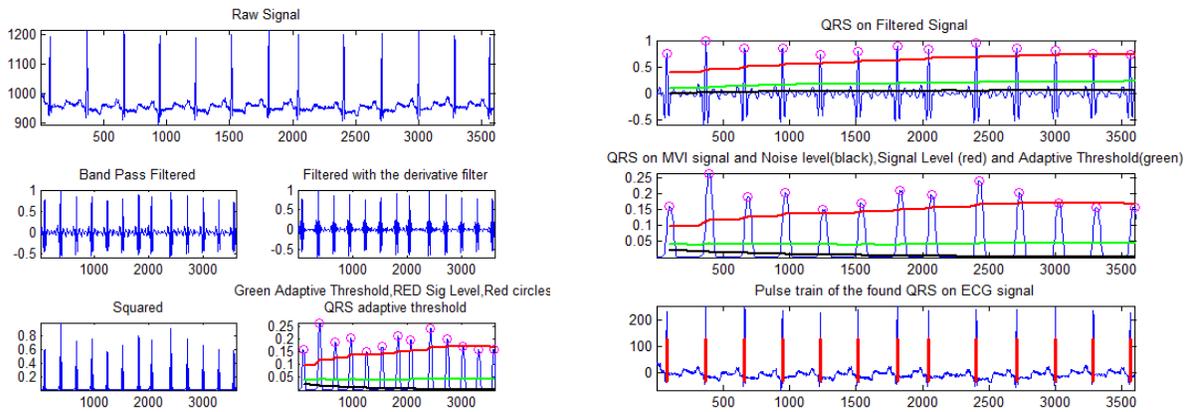

Fig.11. Signal Processing of Arrhythmia Raw ECG signal

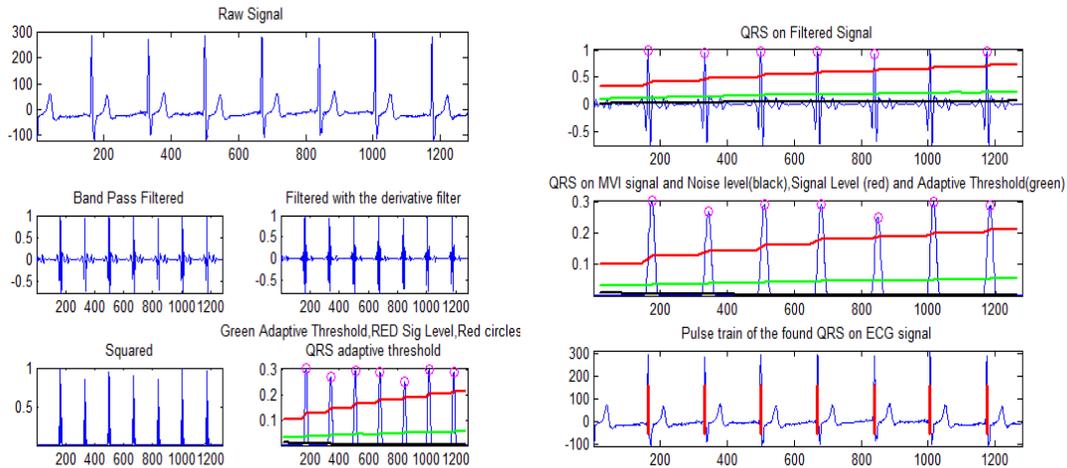

Fig.12. Signal Processing of Supraventricular Arrhythmia Raw ECG signal

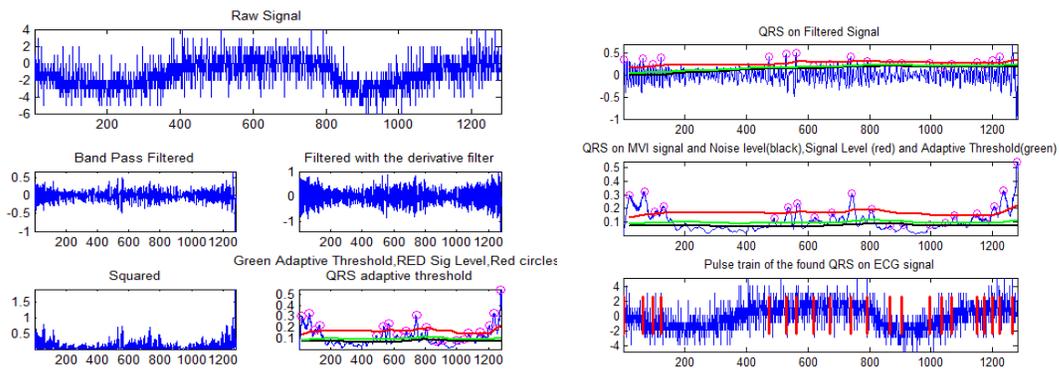

Fig.13. Signal Processing of Atrial fibrillation Raw ECG signal

## A. Artificial Neural Networks classifier

Artificial Neural Networks (ANN) is a computing system which draws parallel from the neurons in the human body and cause changes in the flow path based on the information received. It can be used as a tool for classification, pattern recognition and in other aspects of machine learning. ANN can be used as a powerful tool in order to diagnose diseases and forms an integral part of machine learning which helps in an automated calculation of the disease.

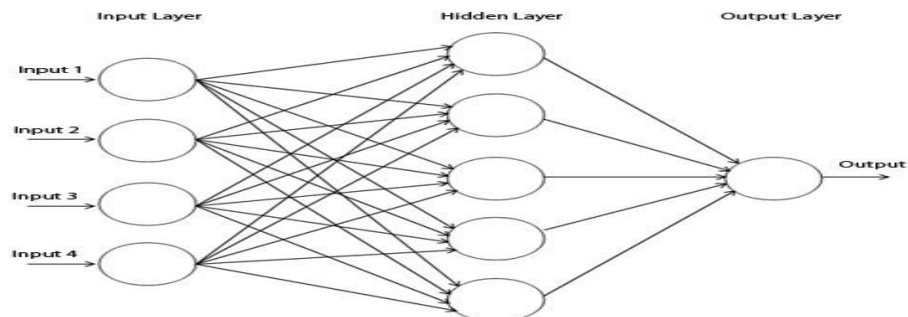

Fig. 14. Standard single-layered ANN structure [4]

We make use of Scaled Conjugate Grading (SCG), which is a supervised machine learning algorithm for feedforward neural networks [7] [8]. Our work involves the use of four neural network- feed forward network, fit network [17] [18], pattern network [19] and cascade forward network [20]. Feed forward network consists of layers of neurons with hidden neurons interconnected to each other. Fit network is a specialized form of Feed forward network [17] which uses Genetic algorithm to govern the learning parameter. Pattern networks are derived from feed forward networks that can be trained to classify a dataset based on target classes [19] and mainly used to identify the pattern in the data. Cascade forward networks are derived from feed forward network and include connection of input nodes and all the previous layer nodes to the following layers [20].The network is responsible for classification of the ECG signal into one of the five possible categories (Normal heart beat or any one of the four diseases).Comparison between all the algorithms used in our work and show the best accuracy obtained.

**Proposed Method**

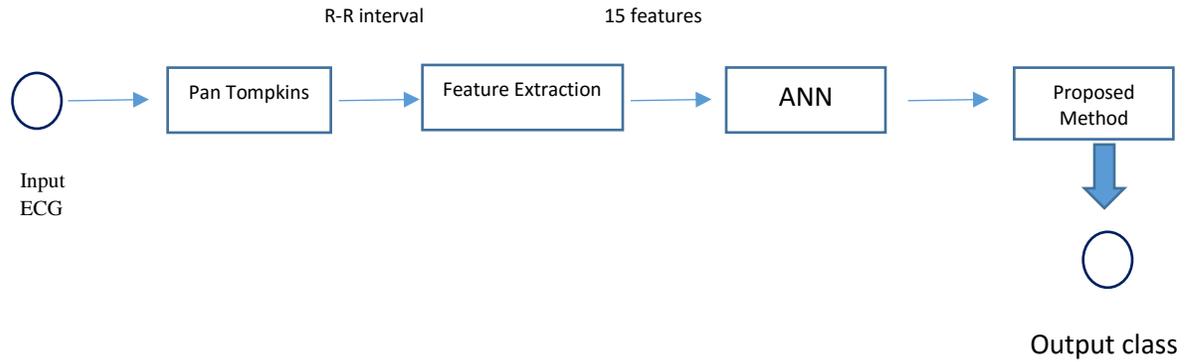

Neural network is trained with 70% of total samples in each case and each sample consist of 15 features by keeping base as RR intervals. In our proposed method we trained 10 network named from Net1-Net10 (Table-1 ).each Net consist of trained features of two diseases and accuracy of each Net case has been projected in table 2(excluding normal case).Since we have four diseases and one normal case hence total combination by combining two cases will be ten and each diseases will repeat four times refer table-1.By using ten Net a condition table(table-1) is formed which clearly explains parameter need to be considered during post classification from NN output .

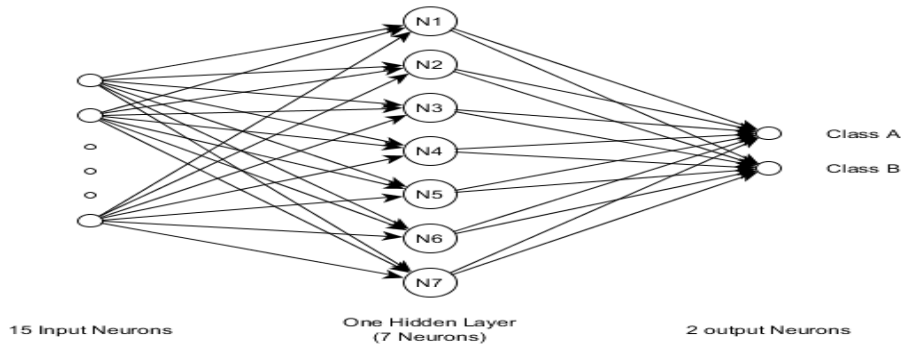

Fig. 15. ANN Structure with 5 input neurons, 7 neurons in 1 hidden layer and 2 output classes

$$X_{train} = \begin{bmatrix} x_{11} & x_{21} & x_{31} & \cdots & x_{1\,15} \\ x_{21} & x_{22} & x_{23} & \cdots & x_{2\,15} \\ \vdots & \vdots & \vdots & \vdots & \vdots \\ x_{n1} & x_{n2} & x_{n3} & \cdots & x_{n\,15} \end{bmatrix} \quad X_{testing} = \begin{bmatrix} x_{11} & x_{21} & x_{31} & \cdots & x_{1\,15} \\ x_{21} & x_{22} & x_{23} & \cdots & x_{2\,15} \\ \vdots & \vdots & \vdots & \vdots & \vdots \\ x_{m1} & x_{m2} & x_{m3} & \cdots & x_{m\,15} \end{bmatrix} \qquad (2)$$

Each ECG signal is used to extract 15 features represented by {$x_1$, x2, x3… x15} used to form the matrices given above with 'n' as the number of samples used for training and 'm' as the number of samples used for validation. From each classifier we get two output which tells about the sample that belong to two classes .If the first (A) output neuron probability is high then it is considered the sample belongs to class 1 (winning class)else ,it is class 2(winning class).In table -1 each Net has two output either the output belongs to A sample or B sample.

In table 1 the last five column mentioned from A-E are the four disease and one normal ECG case. Second column implies the cases taken and first column with Net (trained network varies from 1-10) respectively. The number "0","1" and"2" are used in the table to denote the wining class as explained above .For example in a Net1 case two diseases sampled from A and B are used to train the network then "1" denoted the sample belongs to disease A ,"2" denotes disease B and "0" implies not applicable diseases . In our proposed method we trained 10 combination of the classes and dedicated 10 networks, moreover accuracy is presented in the table IV, with the help of trained networks the following process has been carried out. The condition table is used to find the value of flag variables FA to FE

Table III. Condition Table taking two signal classes at a time

| Networks | | A | B | C | D | E |
|---|---|---|---|---|---|---|
| Net 1 | AB | 1 | 2 | 0 | 0 | 0 |
| Net 2 | AC | 1 | 0 | 2 | 0 | 0 |
| Net 3 | AD | 1 | 0 | 0 | 2 | 0 |
| Net 4 | AE | 1 | 0 | 0 | 0 | 2 |
| Net 5 | BC | 0 | 1 | 2 | 0 | 0 |
| Net 6 | BD | 0 | 1 | 0 | 2 | 0 |
| Net 7 | BE | 0 | 1 | 0 | 0 | 2 |
| Net 8 | CD | 0 | 0 | 1 | 2 | 0 |
| Net 9 | CE | 0 | 0 | 1 | 0 | 2 |
| Net 10 | DE | 0 | 0 | 0 | 1 | 2 |

For coding convenience we use NetX for "1" and NetX' for "2" .either NetX or NetX' will be high according to the condition table and used to determine the value of flag variables FA,FB,FC,FD and FE ,these variable values ranges from 0 to 4.

Flag variables are defined as:

FA = Sum (Net1, Net2, Net3, Net4); FB = Sum (Net1', Net5, Net6, Net); FC = Sum (Net2', Net5', Net8, Net9)

FD = Sum (Net3', Net6', Net8', Net10); FE = Sum (Net4', Net7', Net9', Net10')

Finally, Maximum of FA, FB, FC, FD and FE will take as the output.

**RESULTS**

DATASET

For signal processing (feature extraction), the database of all the required samples was collected from physionet.org. The database of the following diseases was collected from the MIT-BIH database [2] [16]:

- Sleep Apnea
- Normal Sinus Rhythm
- Long term Atrial fibrillation
- Arrhythmia
- Supraventricular Arrhythmia

Each file was downloaded as a Matlab file and used for further processing. 70% of the sample has been taken for training and rest 30% is used for testing

Taking 2 diseases at a time, the ANN were trained and tested with test files of both classes and its accuracy was calculated in table IV. The * denotes the particular case has been given in testing

Table IV. Classification Performance in terms of accuracy with 2 classes at a time, * denotes test file used

| Signal Classes | Cascade Net. | Feed forward Net. | Fit Net. | Pattern Net. |
|---|---|---|---|---|
| *Arrhythmia * ,Long Term AF* | 66.67 | 93.33 | 93.33 | 80 |
| *Arrhythmia ,Long Term AF ** | 96 | 100 | 100 | 96 |
| *Long Term AF * ,Sleep Apnea* | 100 | 100 | 100 | 92 |
| *Long Term AF ,Sleep Apnea ** | 62.5 | 70.83 | 70.83 | 75 |
| *Long Term AF * ,Supraventricular Arrhythmia* | 100 | 100 | 100 | 100 |
| *Long Term AF ,Supraventricular Arrhythmia ** | 100 | 100 | 100 | 100 |
| *Sleep Apnea *, Supraventricular Arrhythmia* | 45.83 | 70.83 | 45.83 | 62.5 |
| *Sleep Apnea, Supraventricular Arrhythmia ** | 93.61 | 100 | 97.87 | 100 |
| *Arrhythmia *, Sleep Apnea* | 66.67 | 73.33 | 46.66 | 13.33 |
| *Arrhythmia , Sleep Apnea ** | 100 | 79.16 | 79.16 | 83.33 |
| *Arrhythmia * Supraventricular Arrhythmia* | <1 | 20 | <1 | 20 |
| *Arrhythmia Supraventricular Arrhythmia ** | 100 | 95.74 | 93.62 | 95.74 |

For more clarity each trained network accuracy has be present in two rows with four column which includes Cascade Net, Feedforward Net, Fit Net, Pattern Net. As we can see in the table.2 maximum accuracy is obtained on training the network with diseases Long Term Af and Superventricular and least is obtained on classes Arrhythmia, Supraventricular Arrhythmia. Among all network normal Feedforward network which uses SCG algorithm yields the best result. Among Arrhythmia and Long Term AF, Feed forward and Fit Net produces the best result, similarly among long term AF and Sleep Apnea ,Feed forward and Fit Net produces the best result but in the case of sleep Apnea and Supraventricular Arrhythmia ,Feed forward and pattern Net produces the best result and in the case of Arrhythmia and sleep apnea ,cascade and Feed forward produces the moderate results .Since the samples from Arrhythmia ,Supraventricular Arrhythmia are similar because of similarity in the disease it can be improved by providing more sample for training moreover DNN can be used to learn features more accurately. If we see in terms of networks, for cascaded Net, Long term AF and supraventricular produces 100% accuracy and Arrhythmia , supraventricular produces <51% accuracy .similarly for other three network Long term AF and supraventricular produces 100% accuracy .

To compare the results obtained through our method we trained those four network using all case samples which results in five classes and results of multiclass network is shown in Table III.(Normal) , it also show the results obtained from proposed method from duel classifier (binary classifier) .Here the results are obtained for all four diseases and the normal case ,as mentioned above we uses the same four type of network to classify the diseases .

Table III. Classification Performance comparison between normal and proposed Classification method with 5 classes take at a time

| Signal Classes | Cascade Net | | Feed Forward Net | | Fit Net | | Pattern Net | |
|---|---|---|---|---|---|---|---|---|
| | Normal | Proposed | Normal | Proposed | Normal | Proposed | Normal | Proposed |
| **Arrhythmia** | 13.33 | 40.909 | <1 | 63.636 | <1 | 54.54 | <1 | 50 |
| **Normal** | <1 | <1 | <1 | 9.09 | <1 | 9.09 | <1 | 9.09 |
| **Long Term AF** | <1 | 88 | 100 | 100 | 100 | 100 | 100 | 100 |
| **Sleep Apnea** | 4.16 | 88 | 33.33 | 40 | 45.83 | 32 | 62.5 | 37.73 |
| **Supra Arrhythmia** | 78.72 | 4 | 91.49 | 60 | 91.48 | 80 | 87.23 | 95.83 |

From table.3 we can infer that most of cases the proposed method produces better results and superior result is found while classifying Long Term AF, cascade net the accuracy has been increased greater than 87%. In the case normal three among four networks accuracy increased by 9.09 % .Similarly accuracy in proposed method in Cascade-Arrhythmia increases 27.579 %, Feed Forward –arrhythmia increases around 63 %, Fit- arrhythmia increases 54.54 %,Pattern net- arrhythmia 50%.

## CONCLUSION

The network was trained with different number of neurons and gave the best result with 7 neurons in one input layer. Feed Forward net and fit net performed best while classifying between arrhythmia with long term AF, and long term AF with sleep apnea. All the 4 networks performed equally well while classifying between long term AF and supraventricular arrhythmia. Feed forward net performed best while classifying between sleep apnea with supraventricular arrhythmia. Cascade net provides best performance when classifying between Arrhythmia and sleep apnea. Table V compares the performance of the normal network multi-classification and the proposed post-classification on binary classification respectively .The proposed method performs better than the multi-classification network in all the cases of Arrhythmia, normal sinus rhythm and long term AF. Moreover it produces superior results than the multi-classification network when used with cascade and feed forward net in case of sleep apnea. It also performs better in case of supraventricular arrhythmia when used with pattern net.Hence, it can be concluded that the proposed method provides a comparatively efficient way to classify ECG signals among the 5 classes taken. It can also be concluded that feed forward net provides the best solution in most cases when comparing the above diseases taken 2 at a time.